\documentclass[letterpaper]{article}
\usepackage{aaai}
\usepackage{times}
\usepackage{helvet}
\usepackage{courier}
\usepackage{tabu}

\title{Identifying Condition-Action Statements in Medical Guidelines Using Domain-Independent Features}
\author{ Hossein Hematialam \and Wlodek Zadrozny \\ Computer Science Department\\
University of North Carolina at Charlotte\\
9201 University City Blvd\\
Charlotte, NC 28223}

\begin{document}
\maketitle

\begin{abstract}
This paper advances the state of the art in text understanding of medical guidelines by releasing two new annotated clinical guidelines datasets, and establishing baselines for using machine learning to extract condition-action pairs. In contrast to prior work that relies on manually created rules, we report experiment with several  supervised machine learning techniques to classify sentences as to whether they express conditions and actions. We show the limitations and possible extensions of this work on text mining of medical guidelines.

\end{abstract}

\section{Introduction}
Clinical decision-support system (CDSS) is any computer system intended to provide decision support for healthcare professionals, and using clinical data or knowledge \cite{musen2014clinical}. The classic problem of diagnosis is only one of the clinical decision problems. Deciding which questions to ask, tests to order, procedures to perform, treatment to indicate, or which  alternative medical care to try, are other examples of clinical decisions. CDSSs generally fall into two categories \cite{musen2014clinical} 
\begin{itemize}
	\item Determining "what is true" about a patient (usually what the correct diagnosis is).
	\item Determining "what to do" for the patient (usually what test to order, whether to treat, or what therapy plan to institute).
\end{itemize}
Most of the questions physicians need to consult about with CDSSs are from the latter category. Medical guidelines (also known as clinical guidelines, clinical protocols or clinical practice guidelines) are most useful at the point of care and answering to "what to do" questions.

Medical guidelines are systematically developed statements to assist with practitioners' and patients' decisions. They establish criteria regarding diagnosis, management, and treatment in specific areas of healthcare. For example, a sentence such as "if the A1C is 7.0\% and a repeat result is 6.8\%, the diagnosis of diabetes is confirmed" in medical guidelines determines what is true about a patient. Sentences such as "Topical and oral decongestants and antihistamines should be avoided in patients with ABRS" guide what to do or not to do with a patient. These examples illustrate conditions, criteria applicable to patients, and consequences of the conditions. The consequences may refer to activities, effects, intentions, or events. If a guideline-based CDSS needs to answer "what to do" questions, it has to have access to condition-action statements describing under what circumstances an action can be performed.
 
Medical guidelines contain many condition-action statements. Condition-action statements provide information about expected process flow. If a guideline-based CDSS could extract and formalize these statements, it could help practitioners in the decision-making process. For example, it could help automatically asses the relationship between therapies, guidelines and outcomes, and in particular could help the impact of changing guidelines.

However, completely automated extraction of condition-action statements does not seem possible. This is due among other things to the variety of linguistic expressions used in condition-action sentences. For example, they are not always in form of "\{if\} condition \{then\} action''. In the sentence "Conditions that affect erythrocyte turnover and hemoglobin variants must be considered, particularly when the A1C result does not correlate with the patient's clinical situation'', we have a condition-action sentence without an "\{if\}" term.

We propose a supervised machine learning model classifying sentences as to whether they express a condition or not. After we determine a sentence contain a condition, we use natural language processing and information extraction tools to extract conditions and resulting activities. 

With the help of a domain expert, we annotated three sets of guidelines to create gold standards to measure the performance of our condition-action extracting models. The sets of guidelines are: hypertension \cite{doi:10.1001/jama.2013.284427}, chapter4 of asthma \cite{british2008british}, and rhinosinusitis \cite{chow2012idsa}. Chapter 4 of asthma guidelines was selected for comparison with prior work of Wenzina and Kaiser \cite{wenzina2013identifying}. We have annotated the guidelines for the conditions, consequences, modifiers of conditions, and type of consequences.
These annotate sets of guidelines are available for experiments https://www.dropbox.com/.

%

\section{Related Work}

We will briefly discuss the modeling and annotation of condition-action for medical usage in this section. Our corpus and method of identifying conditions in clinical guidelines is explained in section 3. 

Research on CIGs started about 20 years ago and became more popular in the late-1990s and early 2000s. Different approaches have been developed to represent and execute clinical guidelines over patient-specific clinical data. They include document-centric models, decision trees and probabilistic models, and "Task-Network Models"(TNMs) \cite{peleg2003comparing}, which represent guideline knowledge in hierarchical structures containing networks of clinical actions and decisions that unfold over time. 
Serban et. al \cite{serban2007extraction} developed a methodology for extracting and using linguistic patterns in guideline formalization, to aid the human modellers in guideline formalization and reduce the human modelling effort. 
Kaiser et. al \cite{kaiser2011identifying} developed a method to identify activities to be performed during a treatment which are described in a guideline document. They used relations of the UMLS Semantic Network \cite{mccray1989umls} to identify these activities in a guideline document. Wenzina and Kaiser \cite{wenzina2013identifying} developed a rule-based method to automatically identifying conditional activities in guideline documents.They achieved a recall of 75\% and a precision of 88\% on chapter 4 of asthma guidelines which was mentioned before.

\section{Condition-Action Extraction}
Medical guidelines’ condition-action statements provide information to determine "what to do" with a patient. Other types of consequences of a condition in a sentence may help practitioner to find what is true about a patient. In this paper, we propose an automated process to find and extract condition-action statements from medical guidelines. We employed NLP tools and concepts in the process to achieve more general models.

We define the task as classification task. Given an input statement, classify it to one of the three categories: NC (no condition) if the statement doesn’t have a condition; CA if the statement is a condition-action sentence; and CC (condition-consequence) if the statement has a condition which has a non-action consequence. For a CDSS, to determine both "what is true" about a patient and "what to do" with a patient, CC and CA statements can be merged to one category. 

There are limitations in this specification of classification categories. For example, guidelines may contain statements with a condition referring to a consequence in another statement. Or, we can see condition and effect in two different sentences: "However, there are some cases for which the results for black persons were different from the results for the general population (question 3, evidence statements 2, 10, and 17). In those cases, separate evidence statements were developed." 

In this work we focus only on statements that follow the above sentence categorization rules. This allows us to make clear comparison to prior work e.g. by Wenzina and Kaiser \cite{wenzina2013identifying}. They annotated chapter 4 of asthma and other guidelines. They used information extraction rules and semantic pattern rules to extract conditional activities, condition-action statements. We use POS tags as features in the classification models. In our opinion, using POS tags instead of semantic pattern rules makes our model more domain-independent, and therefore more suitable for establishing baselines, not only for text mining of medical guidelines but also in other domains, such as text mining of business rules. But we also expect to improve the performance of our extraction programs by adding semantic and discourse information (this work is ongoing).
%
%

%
%
%

\subsection*{Classification}
Most of the condition-action sentences have a modifier in the sentences. For example, in "In the population aged 18 years or older with CKD and hypertension, initial (or add-on) antihypertensive treatment should include an ACEI or ARB to improve kidney outcomes", we have "the population aged 18 years or older with CKD and hypertension" as a condition and "\{in\}" is the modifier. "If", "in", "for", "to", "which", and "when" are the most frequent modifiers in our guidelines. 

We used CoreNLP \cite{manning-EtAl:2014:P14-5} Shift-Reduce Constituency Parser to parse sentences in guidelines. As we mentioned, "if", "in", "for", "to", "which", and "when" are the most frequent modifiers in our guidelines. "If", "in", and "for" are tagged as "IN" which represents preposition or subordinating conjunction. "To" is tagged as "TO" and "when" and "which" are tagged as "WHADV".  We used regular expressions to find those parses which are promising candidates for extraction of condition-action pairs; for example, we selected sentences which include these tags: IN, TO and WHADVP.

We extracted part of speech (POS) tags as our features for our model. Each candidate sentence has at least one candidate condition part. We extract these parts by regular expressions. Each part of sentence which starts with below patterns is a candidate condition part:
\begin{itemize}
	\item "\textbackslash((SBAR|PP) \textbackslash(IN"
	\item "\textbackslash(SBAR \textbackslash(WHADVP"
	\item "\textbackslash(PP \textbackslash(TO"
\end{itemize}	

For example, "(ROOT (S (PP (IN In) (NP (NP (NNS adults)) (PP (IN with) (NP (NN hypertension))))) (, ,) (VP (VBZ does) (S (VP (VBG initiating) (S (NP (NP (JJ antihypertensive) (JJ pharmacologic) (NN therapy)) (PP (IN at) (NP (JJ specific) (NN BP) (NNS thresholds)))) (VP (VBP improve) (NP (NN health) (NNS outcomes))))))) (. ?)))" is the constituent parsed tree of "In adults with hypertension, does initiating antihypertensive pharmacologic therapy at specific BP thresholds improve health outcomes?". "(PP (IN In) (NP (NP (NNS adults)) (PP (IN with) (NP (NN hypertension)))))" and "(PP (IN at) (NP (JJ specific) (NN BP) (NNS thresholds)))" are two candidate condition parts in this example.

We created features for our model based on POS tags and their combinations. The sets of features and the combinations are learned automatically from annotated examples. We used these novel features to make our model more domain-independent. 


For each sentence, we extracted POS tags, sequences of 3 POS tags, and combination of all POS tags of candidate conditions as features. For example, "PP IN NP NP NNS PP IN NP NN PPINNP INNPNP NPNPNNS NPNNSPP NNSPPIN PPINNP INNPNN PPINNPNPNNSPPINNPNN PP IN NP NN PPINNP INNPNN PPINNPNN PP IN NP JJ NN NNS PPINNP INNPJJ NPJJNN JJNNNNS PPINNPJJNNNNS" represents "In adults with hypertension, does initiating antihypertensive pharmacologic therapy at specific BP thresholds improve health outcomes?" in our model. Note that the glued together part of speech tags are not a formatting error but features automatically derived by our model (from consecutive part of speech tags). 

\section{Evaluation}
\subsection*{Gold Standard Datasets}

We use three medical guidelines documents to create gold standard datasets. They provide statements, tables, and figures about hypertension, rhinosinusitis, and asthma. The creation of the gold standard datasets is described below in detail.

Our data preparation process proceeded as follows: We started by converting the guidelines from PDF or html to text format, editing sentences only to manage conversion errors, the majority of which were bullet points. Tables and some figures pose a problem, and we are simply treating them as unstructured text. We are not dealing at this time with the ambiguities introduced by this approach; we do have plans to address it in future work. 

Using regular expressions, as described above, we selected candidate sentences from text files. Note that candidate sentences do not always include a modifier such as "if" or "in". For example, in "Patients on long-term steroid tablets (e.g. longer than three months) or requiring frequent courses of steroid tablets (e.g. three to four per year) will be at risk of systemic side-effects", there is no modifier in the sentence. 

The annotation of the guidelines text (the next step), focused on determining whether there were condition statements in the candidate sentences or not. The instruction to the annotators were to try to paraphrase candidate sentences as sentences with "if condition, then consequence". If the transformed/paraphrased sentence conveyed the same meaning as the original, we considered to be a condition-consequence sentence. Then we we could annotate condition and consequence parts. For example, we paraphrased "Beta-blockers, including eye drops, are contraindicated in patients with asthma" to "If patients have asthma, then beta-blockers, including eye drops, are contraindicated". The paraphrased sentence conveys same meaning. So it became a condition-consequence sentence in our dataset.  On the other hand, for example, we cannot paraphrase "Further, the diagnostic criteria for CKD do not consider age-related decline in kidney function as reflected in estimated GFR" to an if-then sentence. 

We also annotated the type of sentences based on their semantics: We classified them into three classes: condition-action, condition-consequence(effect, intention, and event) and action. Examples are shown in table 1.
\begin{table}
\begin{center}
 \begin{tabu} to 0.5\textwidth { | X[c] | X[c]|} 
 \hline
Type & Example \\
\hline
Condition-Action & Timely referral is indicated if chronic or recurrent symptoms severely affect the patient's productivity or quality of life.\\
\hline
Condition-effect & Most patients with uncomplicated viral URIs do not have fever. \\
\hline
Action & Adjustment is necessary for fluticasone and mometasone and may also be necessary for alternative devices.\\
\hline
\end{tabu}
\caption{Examples of classified sentence classes}
\label{table:1}
\end{center}
\end{table}

Each sentence was annotated by one domain expert and us (and the disagreements where less than 10 percent). Table \ref{table:2} shows the statistics of the annotated sentences for 3 different medical guidelines. 
\begin{table}
\begin{center}
  \begin{tabu} to 0.5\textwidth { |X[c]|X[c]|X[c]|X[c]|X[c]|} 
\hline
 & Condition-Action & Condition-Effect & Action & No Condition\\
\hline
Asthma & 38 & 7 & 8 & 224\\
\hline
Rhinosinusitis & 97 & 39 & 15 & 726\\
\hline
Hypertension & 63 & 14 & 1 & 238\\
\hline
\end{tabu}
\caption{Statistical information about annotated guidelines}
\label{table:2}
\end{center}
\end{table}

\subsection*{Model Performance}
Hypertension, asthma, and rhinosinusitis guidelines and gold standard datasets were applied to evaluate our model. Since two of these annotated corpora are new, our model is establishing a baseline. The asthma corpus was investigated previously by \cite{wenzina2013identifying}. 

We extracted candidate statements by applying aforementioned regex on POS tags. Hypertension, asthma, and rhinosinusitis guidelines had 278, 172, and 761 candidate statements respectively. By applying this filtering subtask, we get rid of 38, 116, and 5 no condition statement respectively from guidelines. 
We used Weka \cite{hall2009weka} classifiers to create our models. ZeroR, Naïve Bayes, J48, and random forest classifiers were applied in our project. Table \ref{table:3}, \ref{table:4}, and \ref{table:5} show the results of classifiers for each guidelines.The results are based on 10-fold cross-validation on respective datasets. 

\begin{table}[h!]
\begin{center}
 \begin{tabu}to 0.5\textwidth{|c|c|c|c|c|} 
 \hline
Asthma & \multicolumn{3}{|c|}{Condition-Action} & Total\\
\hline
 & Precision & Recall & F-measure & Precision\\
 \hline
 ZeroR& 0 & 0 & 0 & 0.69 \\
 \hline
 NaiveBayes & 0.455 & 0.263 & 0.333 & 0.69\\
 \hline
 J48 & 0.444 & 0.211 & 0.286 & 0.67\\
 \hline
 RandomForest & 0.5 & 0.079 & 0.136 & 0.67\\
 \hline
 \end{tabu}
\caption{Classification results on asthma guidelines. (The ZeroR gives 0 precision and recall, because the majority of the guidelines sentences do not contain conditions and actions). }
\label{table:3}
\end{center}
\end{table}
\begin{table}[h!]
\begin{center}
 \begin{tabu}to 0.5\textwidth{|c|c|c|c|c|} 
 \hline
Rhinosinusitis &\multicolumn{3}{|c|}{Condition-Action} & Total\\
\hline
 & Precision & Recall & F-measure & Precision\\
 \hline
ZeroR & 0 & 0 & 0 & 0.80\\
 \hline
NaiveBayes & 0.5 & 0.412 & 0.452 & 0.80\\
 \hline
J48 & 0.581 & 0.258 & 0.357 & 0.81\\
 \hline
RandomForest & 0.844 & 0.392 & 0.535 & 0.84\\
 \hline
 \end{tabu}
\caption{Classification results on rhinosinusitis guidelines}
\label{table:4}
\end{center}
\end{table}
\begin{table}[h!]
\begin{center}
 \begin{tabu}to 0.5\textwidth{|c|c|c|c|c|} 
 \hline
Hypertension &\multicolumn{3}{|c|}{Condition-Action} & Total\\
\hline
 & Precision & Recall & F-measure & Precision\\
 \hline
 ZeroR & 0 & 0 & 0 & 0.72\\
 \hline
NaiveBayes & 0.581 & 0.397 & 0.472 & 0.74\\
 \hline
J48 & 0.619 & 0.413 & 0.495 & 0.74\\
 \hline
RandomForest & 0.931 & 0.429 & 0.587 & 0.81\\
 \hline
 \end{tabu}
\caption{Classification results on hypertension guidelines}
\label{table:5}
\end{center}
\end{table}
The results show that generally random forest classifier seems to work best in extracting Condition-Action statements. 

Notice that these results are lower than previously reported by \cite{wenzina2013identifying}. The difference is due to our using of completely automated feature selection when training on an annotated corpus, and not relying on manually created extraction rules. In addition, their results demonstrate recalls on activities with specific patterns. If we consider all activities in their annotated corpus, their recall will be 56\%. And if we apply their approach on our annotated corpus, the recall will be 39\%. In ongoing work we hope to reduce or close this gap by adding semantic and discourse information to our feature sets.

\section{Conclusions and Future Work}


We investigated the problem of automated extraction of condition-action from clinical guidelines based on an annotated corpus. We proposed a simple supervised model which classifies statements based on combinations of part of speech tags used as features. We showed results of classifiers using this model on three different annotated datasets which we created. We release these dataset for others to use. 

Obviously, this is very preliminary work. Our work established baselines for automated extraction of condition-action rules from medical guidelines, but its performance is still inferior to a collection of manually created extraction rules. To close this gap we are currently augmenting our model with semantic information along the lines of \cite{kaiser2011identifying} and \cite{wenzina2013identifying}.  In addition, we are beginning to experiment with some discourse relations -- these are important, for example, in understanding of lists and tables.  We also plan to make our annotated datasets more convenient to use by re-annotating them with standard annotation tools e.g. BRAT \cite{stenetorp2012brat}.


\bibliographystyle{aaai}
\bibliography{guidelines}
\end{document}